\documentclass{article}

\usepackage{microtype}
\usepackage{graphicx}
\usepackage{subfigure}
\usepackage{booktabs} 
\usepackage{bbm}
\usepackage{listings}
\usepackage{color}
\usepackage{multicol}
\usepackage{colortbl}
\usepackage{xcolor}
\usepackage{multirow}
\definecolor{darkgreen}{rgb}{0,0.5,0}
\usepackage{hyperref}
\usepackage{amsmath}

\newcommand{\name}{Cog Attention}
\newcommand{\modelname}{Cogformer}
\usepackage[accepted]{icml2025}

\icmltitlerunning{More Expressive Attention with Negative Weights}

\begin{document}

\twocolumn[
\icmltitle{More Expressive Attention with Negative Weights}




\begin{icmlauthorlist}
\icmlauthor{Ang Lv}{ruc,tencent}
\icmlauthor{Ruobing Xie}{tencent}
\icmlauthor{Shuaipeng Li}{tencent}
\icmlauthor{Jiayi Liao}{tencent,ustc}
\icmlauthor{Xingwu Sun}{tencent}
\icmlauthor{Zhanhui Kang}{tencent}
\icmlauthor{Di Wang}{tencent}
\icmlauthor{Rui Yan}{ruc}
\end{icmlauthorlist}

\icmlaffiliation{ruc}{Renmin University of China}
\icmlaffiliation{tencent}{Machine Learning Platform Department, Tencent}
\icmlaffiliation{ustc}{University of Science and Technology of China}

\icmlcorrespondingauthor{Ruobing Xie}{xrbsnowing@163.com}
\icmlcorrespondingauthor{Rui Yan}{ruiyan@ruc.edu.cn}

\icmlkeywords{Attention mechanism, Language models, Negative attention weights}

\vskip 0.3in
]



\printAffiliationsAndNotice{}  

\begin{abstract}
We propose a novel attention mechanism, named \name, that enables attention weights to be negative for enhanced expressiveness.
This stems from two key factors: 
(1) \name\ enhances parameter flexibility. 
For example, unlike traditional softmax attention heads, which use a static output-value (OV) matrix to delete or copy inputs that the heads attend to, \name\ naturally learns to use the sign of dynamic query-key (QK) inner products to represent these operations. 
This enables \name\ to perform multiple operations simultaneously within a single head. 
Meanwhile, \name's OV matrix can focus more on refinement.
(2) \name\ improves the model's robustness against representational collapse by preventing earlier tokens from ``over-squashing'' into later positions.
We develop Transformer-like models that use \name\ as attention modules, including decoder-only models with up to 1 billion parameters for language modeling and U-ViT diffusion models for image generation. 
Experiments show that models using \name\ exhibit superior performance compared to those employing traditional softmax attention modules.
Our approach suggests a promising research direction for rethinking and breaking the entrenched constraints of traditional softmax attention, such as the requirement for non-negative weights.
\end{abstract}

\section{Introduction}
\label{intro}
The Transformer architecture~\cite{NIPS2017_3f5ee243} has achieved success across numerous applications, such as language modeling~\cite{brown2020languagemodelsfewshotlearners} and image generation~\cite{dosovitskiy2021an}.
A key factor in its success is the softmax attention mechanism~\cite{bahdanau2016neuralmachinetranslationjointly}.

\begin{figure}
\skip 0.1in
\begin{center}
    \centerline{\includegraphics[width=\linewidth]{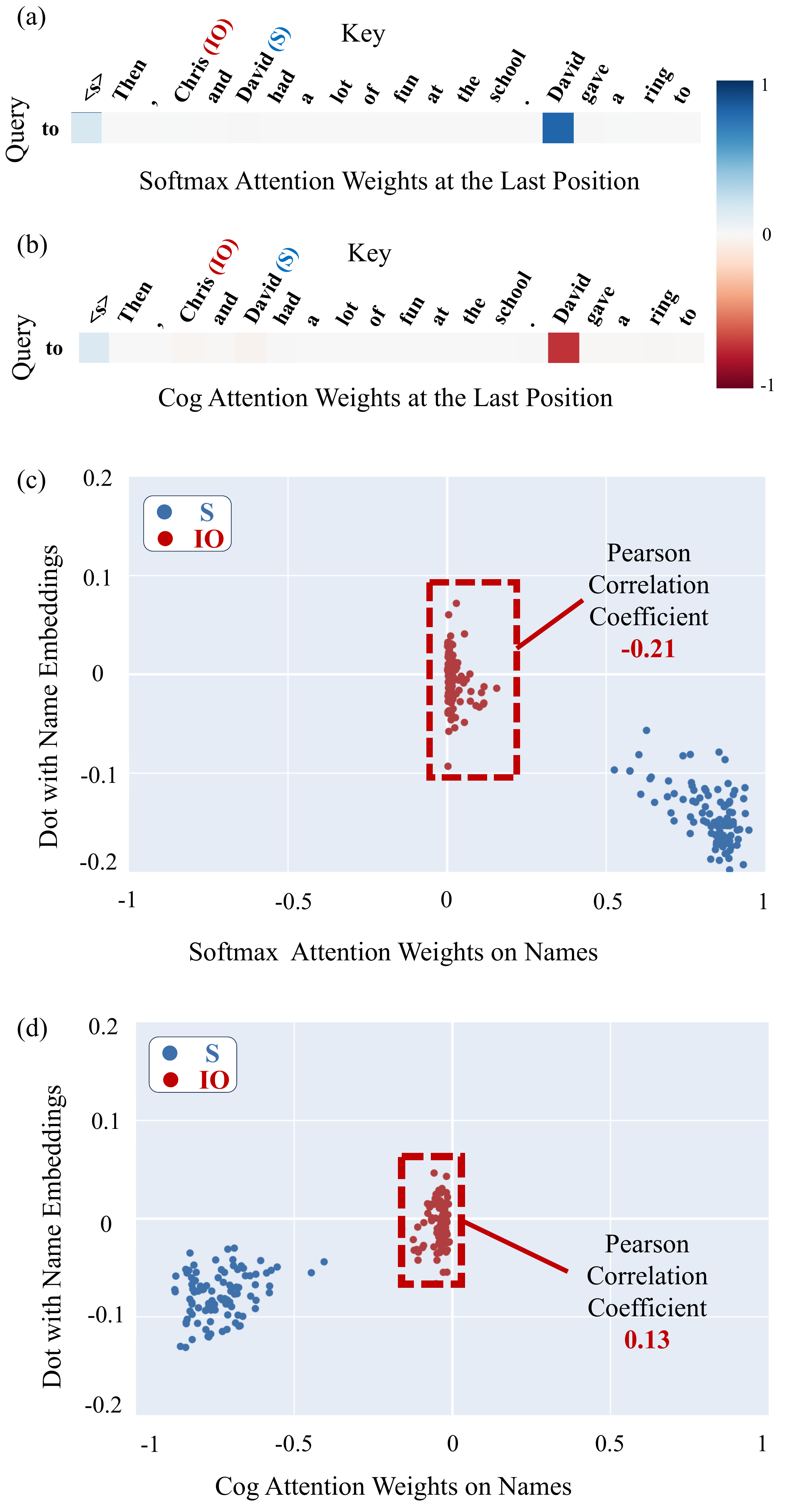}}
    \caption{
In the Indirect Object Identification (IOI) task~\cite{wang2023interpretability}, a language model should identify the indirect object ($IO$) from a context that includes both the $IO$ and a subject ($S$). 
Figures (a) and (b) illustrate how \name\ and softmax attention perform IOI through a process of elimination: 
a softmax attention head with a deletion-function OV matrix eliminates all attended tokens. 
While the $IO$ token receives less attention than $S$, it is also deleted. 
In contrast, \name\ shifts functions like deletion or copying from a static OV matrix to dynamic query-key inner products, allowing the head to assign negative weights to $S$ tokens for elimination while preserving the $IO$s.
Figures (c) and (d) show attention weights on names versus the direction of the heads' output across the entire dataset. 
\name\ preserves the $IO$s better.
For further details, please see Section~\ref{sec:mechanism}.
}
\label{fig:head-compare}
\end{center}
\end{figure}

Softmax ensures non-negative attention weights, but we argue that it limits the expressiveness of the attention mechanism.
Figure~\ref{fig:head-compare} shows one way in which negative attention weights can enhance the model's expressiveness while using the same number of parameters: in a softmax attention head, the query-key (QK) matrix~\cite{elhage2021mathematical} determines the relevant tokens for attention, while the output-value (OV) matrix governs the processing of these attended tokens (e.g., deletion or copying). 
Suppose a softmax attention head has an OV matrix capable of deleting tokens attended to by the QK matrix; since attention weights must be non-positive, a useful token would also be somewhat deleted. 
By allowing negative attention weights, however, deletion or copying can be expressed through the sign of the attention weight and accomplished during the weighted summation of value vectors. 
This functional shift also allows the OV matrix to focus more on higher-level tasks, such as refinement or modification, rather than solely handling contextual deletions or copies.
Consequently, the risk of ``friendly fire'' on useful tokens is mitigated.

Despite the potential benefits of incorporating negative weights in attention mechanisms, this question has been rarely explored. 
Apart from the common belief that attention weights should naturally be non-negative, introducing negative weights can lead to challenges such as training instability, numerical overflow, and difficulties in attention normalization due to issues like division by zero.

In this paper, we propose a novel attention mechanism named \name\footnote{The name is derived from the attention pattern, which resembles cogs (see Figure~\ref{fig:attn-my}).} that enables negative weights. 
\name\ exhibits superior properties in terms of its working mechanisms and outperforms softmax attention in various applications, without introducing additional parameters or hyperparameters.
In Section~\ref{sec:mechanism}, we provide mechanistic interpretation of its enhanced expressiveness: 

(1) We identify several \name\ heads in our pre-trained language models that share the same working mechanism as exemplified above (Figures~\ref{fig:head-compare}(b) and (d)), which shift the contextual process from the static OV matrix to dynamic QK inner products, with the OV matrix focusing more on refinement or modification. 
Irrelevant tokens are assigned negative weights for elimination, while other tokens are preserved.
This demonstrates \name's enhanced flexibility and expressiveness compared to softmax attention.

(2) We empirically and theoretically demonstrate that models using \name\ exhibit improved robustness against representational collapse~\cite{liu-etal-2020-understanding,xie2023residualtransformerdualresidual}.
Representational collapse refers to the phenomenon where representations become homogeneous, especially in the later positions of a sequence, within deep Transformer models.
\citet{barbero2024transformers} contended that this issue arises because earlier tokens are ``over-squashed'' into later positions as the layer depth increases.
The negative weights in \name\ reduce the effective information paths from earlier tokens to later positions, thereby alleviating over-squashing and, consequently, mitigating representational collapse.

In Section~\ref{sec:exp}, we develop Transformer-like models that use \name\ as attention modules.
Specifically, we train decoder-only language models with parameter scales ranging from 140M to 1B for language modeling, and U-ViT diffusion models~\cite{bao2022all} for both unconditional and text-conditioned image generation. 
Our results show that, across a wide range of tasks, these models equipped with \name\ achieve improved performance compared to the vanilla Transformer architecture using softmax attention.
Additionally, in Section~\ref{sec:discuss}, we discuss some important properties of \name\ that could be beneficial for future work.

\section{Method}

We begin by presenting the formulation of our proposed \name, followed by a discussion of the design motivation and underlying principles, as well as how to efficiently implement it.

\subsection{Formulation}
\label{sec:formulation}
Let $\textbf{q}, \textbf{k}, \textbf{v} \in \mathbbm{R}^{T \times d}$ represent the query, key, and value vectors in an attention head. 
$T$ is the number of input tokens, and $d$ is the dimension of the hidden states.
The general attention computation can be expressed as follows:
\begin{equation}
    \begin{aligned}
        \textbf{p}_i &= \textbf{q}_{i}\textbf{k}^\top,\\
        \textbf{a}_i &= \phi(\textbf{p}_i),\\
        \textbf{o}_i &= \sum^{i}_{j=0} \textbf{a}_{i,j} \textbf{v}_{j}, \\ 
    \end{aligned}
    \label{eq:attn}
\end{equation}
where $\textbf{p}_i\in \mathbbm{R}^{1 \times T}$ is the $i$-th row of the inner-product matrix.
$\textbf{a}_i \in \mathbbm{R}^{1 \times T}$ is the $i$-th row of the attention weights.
$\textbf{o}_i \in \mathbbm{R}^{1 \times d}$ is the weighted sum of attended vectors.\footnote{Eq.\ref{eq:attn} is a causal attention formulation, as a token $i$ can only attend to preceding tokens $j \leq i$.}

$\phi(\cdot)$ is \texttt{softmax} function in a traditional attention module:
\begin{equation}
    \begin{aligned}
        \texttt{softmax}(\textbf{p}_{i})_{j} = \frac{\texttt{exp}(\textbf{p}_{i,j} - m_i)}{\sum^{i}_{k=0}\texttt{exp}(\textbf{p}_{i,k} - m_i)},
    \end{aligned}
    \label{eq:softmax-formula}
\end{equation}
where $m_i = \texttt{max($\textbf{p}_i$)}$.
The subtraction of $m_i$ in the exponent aims to avoid numerical overflow.

In \name, we redefine $\phi(\cdot)$ as follows:
\begin{equation}
    \begin{aligned}
    \phi(\textbf{p}_{i})_{j} = \frac{\texttt{SignExp}(\textbf{p}_{i,j})}{\sum^{i}_{k=0} \lvert\texttt{SignExp}(\textbf{p}_{i,k})\rvert},
    \text{ where} \\
    \texttt{SignExp}  (\textbf{p}_{i,j}) = s_{i,j} \cdot \texttt{exp}(s_{i,j} \cdot \textbf{p}_{i,j} - \tilde{m}_i), \\
        \tilde{m}_i = \texttt{max($\lvert\textbf{p}_i\rvert$)}
        \ \ \text{\normalsize{and}} \ \ 
         s_{i,j} = \texttt{sign}(\textbf{p}_{i,j}). \\
    \end{aligned}
    \label{eq:leon-formula}
\end{equation}
This formulation enables negative attention weights.
In the following subsection, we introduce the design motivations and explain the underlying principles.

\subsection{Design Principle}

\textbf{(1) The way to introduce negative weights.}
Although the inner product of query and key vectors naturally contains both positive and negative values, we apply an exponential function to this inner product and subsequently recover the sign of each term.
This method is driven by our observation that an effective attention pattern for convergence must demonstrate sufficient kurtosis—that is, it should be sparse and sharp enough.
Without the exponential function, the attention pattern tends to be too flat, which can impede training convergence.
We also tried using a cubic function as an alternative, which would eliminate the sign recovery process while still offering attention weights with adequate kurtosis.
However, we ultimately chose the exponential function because of its convenience in gradient computation.

\begin{figure}[!t]
\begin{center}
\centerline{\includegraphics[width=\linewidth]{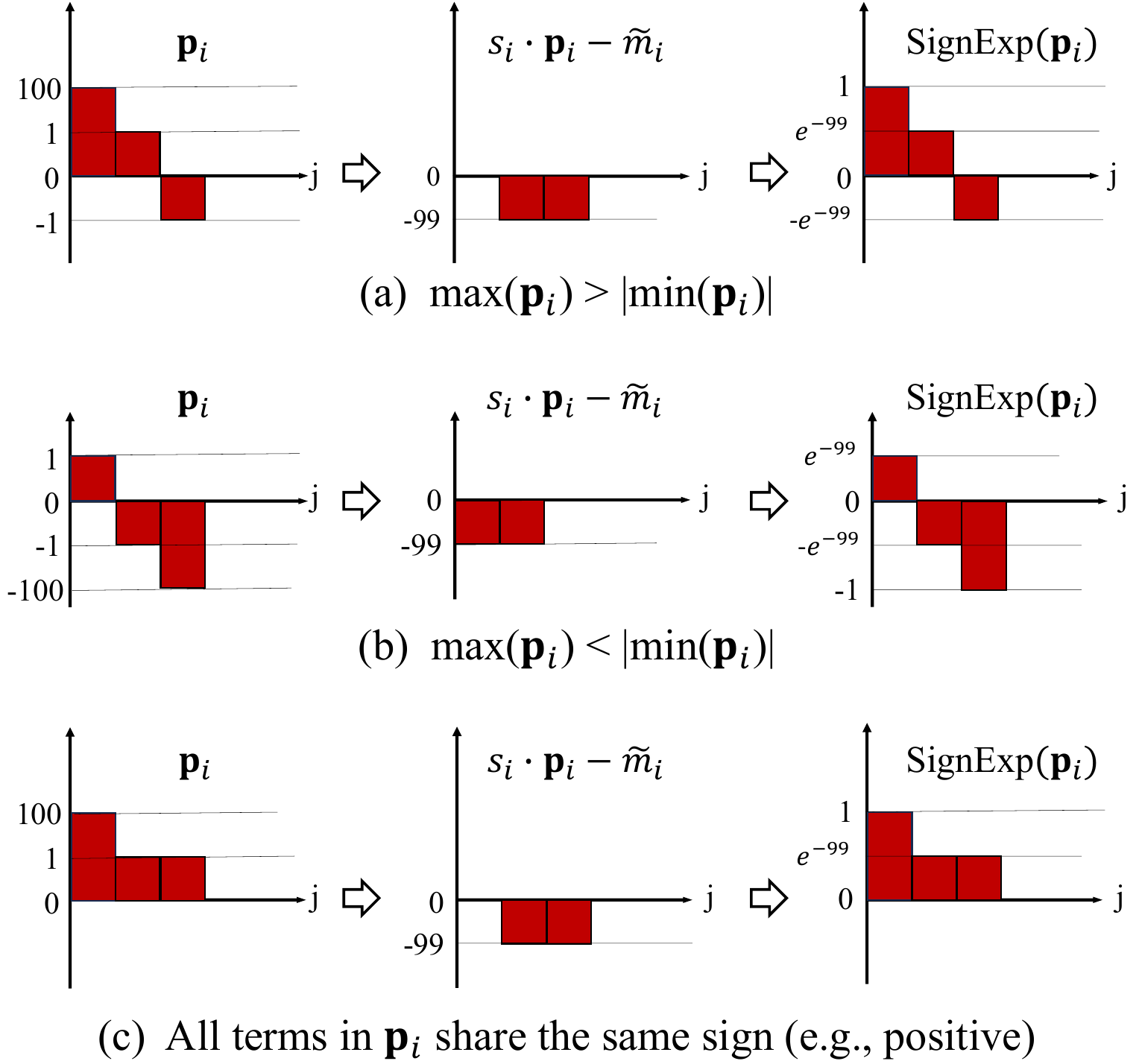}}
\caption{
    The subtraction of the maximum absolute value from a row of query-key inner products avoids numerical overflow.
    Meanwhile, this approach maintains the relative importance of negative and positive inner products in the final attention weights.}
    \label{fig:max}
\end{center}
\vskip -0.15in
\end{figure}

\textbf{(2) The way to avoid numerical overflow.} 
In Eq.\ref{eq:leon-formula}, we avoid numerical overflow in exponential functions by subtracting the maximum \textit{absolute} value from the exponent. 
This differs from softmax, which subtracts the maximum value, as seen in Eq.\ref{eq:softmax-formula}. 
Our approach ensures that the maximum input to $\texttt{SignExp}(\cdot)$ remains 0, effectively avoiding overflow caused by a large $s_{i,j} \cdot \textbf{p}_{i,j}$, as shown in Figure~\ref{fig:max}.

Additionally, subtracting the maximum \textit{absolute} value effectively preserves the relative magnitude between negative and positive inner products in the final attention weights. 
For instance, as shown in Figure~\ref{fig:max}(b), if the minimum inner product is –100 and the maximum is 1, we expect the final attention weights to reflect that the absolute value of the smallest negative weight still exceeds that of the highest positive weight.
Conversely, Figure~\ref{fig:max}(a) illustrates the opposite scenario, where positive weights dominate.

\begin{figure*}[t]
\begin{lstlisting}[language={Python}]
import torch
import torch.nn.functional as F

def Cog_Attention_naive(q, k, v, mask):
    # q, k, and v's shape: T x d
    # mask shape: T x T, 0 for tokens to be masked, 1 for others
    p = q @ k.transpose(1,0)
    p = p * mask
    p_sign = torch.sign(p)
    max_p = torch.abs(p).max(-1, keepdim=True).values
    e = p_sign * torch.exp(p_sign * p - max_p)
    attn_w = e / torch.abs(e).sum(-1, keepdim=True)
    return attn_w @ v

def Cog_Attention_faster(q, k, v, mask):
    p = q @ k.transpose(1,0)
    abs_p = torch.abs(p)
    abs_p.masked_fill_(mask == 0, -float("inf"))
    attn_w = torch.sign(p) * F.softmax(abs_p, dim=-1)
    return attn_w @ v
\end{lstlisting}
\caption{A naive implementation of \name\ in Pytorch, alongside an equivalent yet faster implementation.
By writing a fused kernel in Triton~\cite{10.1145/3315508.3329973}, \name\ achieves the same efficiency as softmax attention.}
\label{fig:code}
\vskip 0.05in
\end{figure*}

\textbf{(3) The way to normalization.}
In the softmax function, the denominator in Eq.~\ref{eq:softmax-formula} is the sum of the numerators, serving to normalize the outputs. 
This process ensures that the resulting attention weights sum to 1, with each term constrained within the range of $[0, 1]$. 
Previous studies suggested that, for better convergence, the sum of attention weights in a row should remain stable, although not necessarily equal to a constant 1~\cite{wortsman2023replacingsoftmaxreluvision}. 
If not, training tricks are necessary to maintain convergence and training stability~\cite{ramapuram2024theoryanalysisbestpractices}.

However, we challenge these beliefs.
As shown in Eq.~\ref{eq:leon-formula}, normalizing by summing all outcomes of \(\texttt{SignExp}(\cdot)\) may result in a zero denominator, causing NaN errors. 
To address this, we propose using the sum of the \textit{absolute} values of the outcomes of \(\texttt{SignExp}(\cdot)\) as the denominator. 
This adjustment leads to a non-constant summation across a row in the attention weight matrices. 
Nevertheless, our experiments demonstrate that \name\ maintains training stability and does not hinder convergence. 
This is because: (1) \(\sum_{j=0}^{i} \lvert \phi(\textbf{p}_i)_{j} \rvert\) remains constant at 1; and (2) based on the observation that the expectation of $\textbf{v}$ during pre-training is close to zero, adding or subtracting these value vectors—assumed to follow a multivariate Gaussian distribution—does not disrupt the norm expectation of the results, i.e., \(\textbf{o}_i\) in Eq.~\ref{eq:attn}.

\subsection{Efficiency}
Figure~\ref{fig:code} presents a naive PyTorch implementation of \name, which closely follows the above description.
Furthermore, we offer a more efficient equivalent version in PyTorch.
Compared to softmax attention, this version still requires additional sign and absolute value operations.
In practice, when implementing \name\ with a fused kernel in Triton~\cite{10.1145/3315508.3329973}, the overhead from these extra operations is negligible, and \name\ is as efficient as softmax attention.
In Appendix~\ref{apx:time}, we provide a detailed comparison of speed and memory usage between \name\ and softmax attention.

\section{Enhanced Expressiveness of \name: A Mechanistic Interpretability Perspective}
\label{sec:mechanism}

In this section, we provide mechanistic interpretability evidence to demonstrate that negative attention weights can enhance the expressiveness of neural networks.

\subsection{Enhanced Flexibility of Attention Heads}

Due to unconstrained attention weights, \name\ enables processes such as deletion or copying, shifting from using a static OV matrix to dynamic query-key products.
This capability is a key advantage of \name, as it facilitates concurrent processes within a single head, thereby enhancing the model's flexibility and expressiveness.

We trained a Transformer language model with 141 million parameters and a Transformer-like language model using \name\ of the same size, respectively. 
Details regarding the model training can be found in Section~\ref{sec:exp}.
We studied the working mechanisms of attention heads on the indirect object identification (IOI) task~\cite{wang2023interpretability}, where the model is provided with a context that includes the names of two people.
For instance, given the input ``Christopher and David had a lot of fun at school. David gave a ring to,'' ``Christopher'' is the indirect object ($IO$), while ``David'' is the subject ($S$). 
The correct answer in the IOI task is always the $IO$, which in this case is ``Christopher.''
There are 100 samples in the dataset.

To identify the most influential attention heads contributing to correct predictions in each model, we employed the path patching algorithm~\cite{wang2023interpretability}. 
We identified two significant heads: the 4th \name\ head in Layer 9 ($\text{CH}_{9.4}$) and the 11th softmax head in Layer 9 ($\text{SH}_{9.11}$). 
These heads accomplish the IOI task through a process of elimination. 
Figure~\ref{fig:head-compare}(a) and (b) illustrate the attention weight patterns for these heads.
Additionally, we computed the attention weights from the final token to both the $IO$ and $S$ tokens, plotted against the inner product of the heads' outputs and the individual embeddings of $IO$ and $S$, as shown in Figure~\ref{fig:head-compare}(c) and (d).

The figures demonstrate that the two heads employ different mechanisms for the elimination process:

$\text{SH}_{9.11}$ assigns a large weight to the $S$ tokens, with its OV matrix~\cite{elhage2021mathematical} generating a vector that opposes the representation of $S$'s embedding. 
Given that the function of the OV matrix is determined by the trained parameters, $\text{SH}_{9.11}$ also eliminates $IO$s due to its non-negative attention weights toward them.
This is supported by Pearson correlation coefficient of $-0.21$ between the attention weights and the inner product, suggesting a weak to medium correlation.
As a result, the non-negative nature of the softmax attention weights limits the performance on the IOI task.

In contrast, $\text{CH}_{9.4}$ assigns negative weights to $S$, effectively eliminating it while assigning minimal weights to $IO$s.
Even if an $IO$ is assigned with a negative weight, we contend that the OV matrix in $\text{CH}_{9.4}$ is relieved from deletion to perform post-processing such as refinement or modification, and thus retains $IO$s.
Evidence for this claim comes from the eigenvalue positivity\footnote{Defined in the ``Summarizing OV/QK Matrices'' section in ~\cite{elhage2021mathematical}} for the OV matrices in two heads.
The eigenvalue positivity, which indicates the OV matrix's tendency toward copying (close to 1), deletion (close to -1), or abstract post-processing such as refinement or modification (close to 0), is 0.78 for $\text{CH}_{9.4}$ and -0.95 for $\text{SH}_{9.11}$, with the absolute value of $\text{CH}_{9.4}$'s eigenvalue positivity being smaller, indicating fewer contextual operations.
Additionally, the low Pearson correlation coefficient of $0.13$ between the attention weights and the inner product suggests little to no correlation between these two variables; that is, $IO$s are hardly eliminated by $\text{CH}_{9.4}$.

\begin{figure}[t]
    \begin{center}
\centerline{\includegraphics[width=\linewidth]{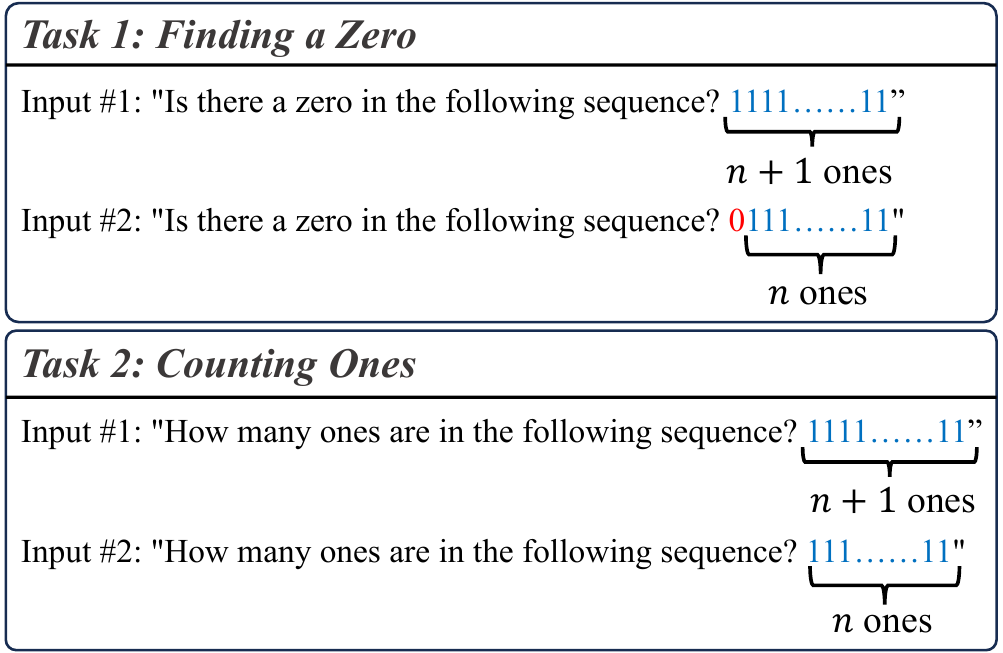}}
    \caption{Two tasks for evaluating the extent of representational collapse in language models.}
    \label{fig:oversquash-task}
\end{center}
\vskip -0.3in
\end{figure}

\begin{figure}[t]
    \begin{center}
\centerline{\includegraphics[width=\linewidth]{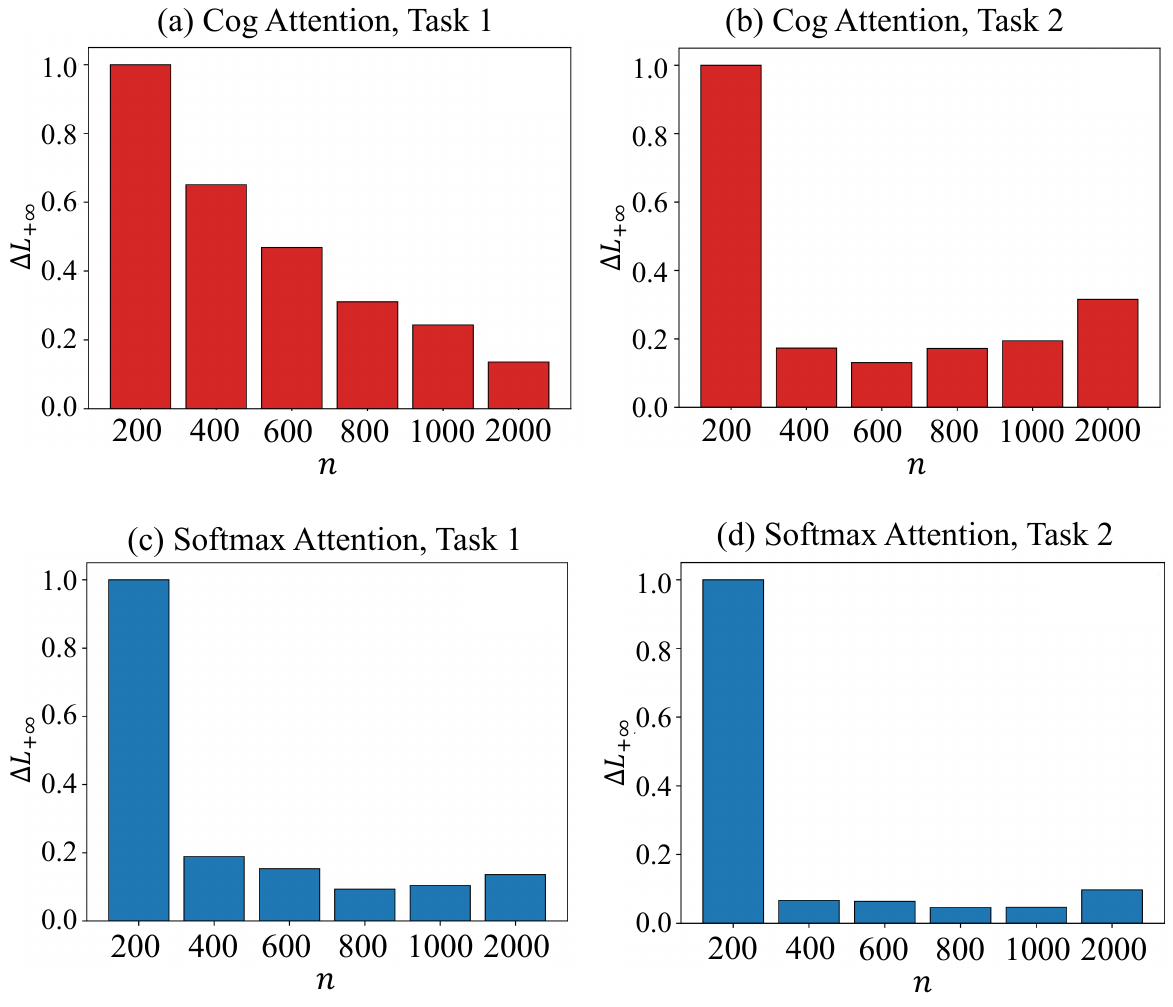}}
\caption{
The output representation difference are measured for Transformer language models utilizing \name\ (subfigures (a) and (b)) and softmax attention subfigures (c) and (d)), respectively. 
\name\ enhances the robustness of language models against representational collapse.}
    \label{fig:oversquash-result}
\end{center}
\vskip -0.2in
\end{figure}

\subsection{More Robustness to Representational Collapse}
\label{sec:rep-exp}
Transformer models suffer from the issue of representational collapse.
Initially, \citet{liu-etal-2020-understanding} and \citet{xie2023residualtransformerdualresidual} described representational collapse as the high similarity of representations between consecutive layers. They attributed this issue to normalization layers and optimizers.
Later, \citet{barbero2024transformers} broadened the definition, identifying another form of representational collapse that occurs within the same layer, where the representations of tokens at different positions become increasingly similar as model depth increases. 
They explain this by suggesting that earlier tokens have more information pathways to the final position than later tokens, leading to the ``over-squashing'' of information from earlier tokens into the final representation. 
This over-squashing causes representational collapse, making it difficult for Transformer models to distinguish between contexts that differ only slightly.
Since the latter form of representational collapse is attributed to the attention mechanism—central to the focus of this paper—we adopt the definition of representational collapse as presented in \cite{barbero2024transformers}.

To evaluate representational collapse in Transformer-based language models, we employ two tasks, as shown in Figure~\ref{fig:oversquash-task}.
Task 1 ``Finding a Zero'' involves processing two input sequences. 
The first sequence consists of \( n + 1 \) ones, while the second begins with a zero followed by \( n \) ones, where \( n \) varies. 
Given an \( n \), let \( \mathbf{y}^{n}_{1} \) and \( \mathbf{y}^{n}_{2} \) represent the output vectors at the final position of the last layer for the first and second inputs, respectively. 
A higher value of \( L_{+\infty} \)-norm of their difference indicates that the model is better at distinguishing between the two similar contexts, reflecting reduced representational collapse~\cite{barbero2024transformers}. 
In this paper, we report the relative \( L_{+\infty} \)-norm, defined as:
\[
\Delta L_{+\infty} = \frac{\lVert \mathbf{y}^{n}_{2} - \mathbf{y}^{n}_{1} \rVert_{+\infty}}{\lVert \mathbf{y}^{200}_{2} - \mathbf{y}^{200}_{1} \rVert_{+\infty}},
\]
because norms are not directly comparable across different models. 
Task 2, ``Counting Ones,'' employs the same evaluation approach.

We evaluate our models on these two tasks.
As shown in Figure~\ref{fig:oversquash-result}, for each \( n \) and across both tasks, models employing \name\ demonstrate greater robustness against representational collapse.
This advantages may be beneficial for tasks involving long and complex contexts, such as retrieval-augmented generation (RAG; \citealp{lewis2020retrieval}).

\textit{Remark.} This enhanced robustness against representational collapse is attributed to the negative weights in \name\, which reduce redundant information pathways from earlier tokens to later positions, thereby mitigating over-squashing. A formal discussion is provided in Appendix~\ref{apx:proof}.

\section{Experiments}
\label{sec:exp}

We construct Transformer-like models using \name\ as the attention module, named \modelname. 
We implement \modelname\ in various tasks, including language modeling and image generation, to serve as a decoder-only model and a diffusion model, respectively. 

\subsection{Decoder-Only Models for Language Modeling}

\textbf{Baselines.}
To show the effectiveness of \name\ beyond the vanilla Transformer, we compare \modelname\ with several Transformer variants that might produce negative attention weights as a by-product of their design, though none are specifically tailored for this purpose.

(1) Differential attention~\cite{ye2024differentialtransformer} assumes attention calculations are noisy and aims to denoise them by subtracting outputs from two attention heads. 
This occasionally produces negative attention weights.

However, this method requires complex initialization and learnable coefficients to balance the two heads, along with additional normalization layers, for stability.

(2) Centered attention~\cite{ali2023centeredselfattentionlayers} attributes the representational collapse to the eigenvalues of the attention matrix being constrained within a limited interval.
It mitigates this issue by shifting the row-wise attention sum from 1 to 0, which may lead to negative weights:
\begin{align*}
    \textbf{o}_i = \sum^{i}_{j=0} (\textbf{a}_{i,j} - \frac{1}{i}) \cdot \textbf{v}_j.
\end{align*}
However, this shift depends on the input length rather than the input content, and does not fully remove the value constraints within the attention matrix, making the approach suboptimal.
Notably, this work can provide additional theoretical insights into the benefits of \name.

\textbf{General Implementation and Hyper-parameters.}
We train decoder-only Transformer language models and their variants using \name\ and its baselines on the RedPajama dataset~\cite{together2023redpajama}. 
Apart from differences in the attention modules, the overall architecture and training hyperparameters remain consistent across all models.

Following the Llama~\cite{touvron2023llamaopenefficientfoundation}, we employ rotary position embedding~\cite{su2023roformerenhancedtransformerrotary} and SwiGLU activation~\cite{shazeer2020gluvariantsimprovetransformer} in the feed-forward networks (FFN). 
RMSNorm~\cite{zhang2019rootmeansquarelayer} is applied prior to both the attention and FFN modules. 
The Llama tokenizer, with a vocabulary of 32,000 tokens, is used.
The models have 141 million parameters, comprising 12 layers, each with 12 attention heads. 
The hidden state dimension is 768, and the intermediate dimension of the MLP layers is 3,072. 
The training context length is 2,048 tokens.

We use a batch size of 64, an initial learning rate of 2e-4, and linear warm-up for the first 2,000 steps followed by a cosine decay schedule to 4\% of the peak learning rate~\cite{StableLM-3B-4E1T}.
The AdamW optimizer~\cite{adamw} is configured with $(\beta_1, \beta_2) = (0.9, 0.95)$, a norm clipping value of 1, and a weight decay of 0.1.
Each model is trained on 100 billion tokens, using 8$\times$A800-80G GPUs for approximately one week.

\textbf{Model-Specific Settings. }
Differential attention requires a complex initialization of relative scales between attention heads. 
We follow the configuration described in the original paper; for details, please refer to~\cite{ye2024differentialtransformer}.  
For \modelname, we preserve softmax attention in both the first and last layers.
This is design choice is explained in Section~\ref{sec:converge}. 
Centered attention does not need special settings.

\begin{table*}[t]
    \caption{Performance comparison across multiple NLP tasks among language models using different attention algorithms. 
    We highlight the top and second-best results in each task with bold text and underlining, respectively. 
    \modelname\ achieves the highest average accuracy compared to other Transformer variants, which may occasionally produce negative attention as a by-product of specific operations.}
        \vskip 0.1in
\begin{center}
\begin{small}
    \label{tab:lm}
    \begin{tabular}{l|c|c|c|c|c|c|c|c}
    \toprule
        Model & ARC-E & ARC-C & PIQA & SIQA &MRPC & SST2 & MNLI & Avg. \\\midrule
        $\text{Transformer}_{\text{141M}}$ & \underline{42.34} & \textbf{19.54} & 57.73 & 37.00 & \underline{60.54} & 51.72 & 32.05 & 42.98 \\
        w/ Differential Attn. & 40.78 & 18.86 & 57.89 & \underline{37.10} & 58.82 & 53.56 & \textbf{34.38} & \underline{43.06} \\
        w/ Centered Attn. & 40.66 & 19.11 & \underline{58.16} & 36.39 & 58.33 & \textbf{55.28} & 32.76 & 42.96 \\\midrule
       $\text{\modelname}_{\text{141M}}$ (Ours) & \textbf{43.90} & \textbf{19.54} & \textbf{59.09} & \textbf{37.36} & \textbf{62.25} & \underline{54.59} & \underline{33.71} &\textbf{44.35}\\\bottomrule
    \end{tabular}
    \end{small}
    \end{center}
    \vskip -0.1in
\end{table*}

\begin{table*}[t]
    \caption{Larger \modelname\ models (600M parameters on 250B tokens and 1B parameters on 50B tokens) outperform Transformer models.}
        \vskip 0.1in
\begin{center}
\begin{small}
    \label{tab:llm}
    \begin{tabular}{l|c|c|c|c|c|c|c|c}
    \toprule
        Model & ARC-E & ARC-C & PIQA & SIQA & MRPC & SST2 & MNLI & Avg. \\\midrule
         $\text{Transformer}_{\text{600M}}$ & 51.01 & 21.33 &  \textbf{64.15} & 40.28 & 58.09 & 54.09 & 33.49 & 46.06 \\
        $\text{\modelname}_{\text{600M}}$ & \textbf{52.40} & \textbf{23.46} & 63.71 & \textbf{40.51} & \textbf{67.16} & \textbf{55.05} & \textbf{33.56}  & \textbf{47.97} \\\midrule
        $\text{Transformer}_{\text{1B}}$ & 47.81 & 20.22 & 61.75 & 37.22 & 58.33 & \textbf{34.81} & 58.49 & 45.52 \\
        $\text{\modelname}_{\text{1B}}$ & \textbf{48.86} & \textbf{21.59} & \textbf{62.24} & \textbf{38.54} & 58.33 & 32.24 & \textbf{61.70} & \textbf{46.21} \\
       \bottomrule
    \end{tabular}
    \end{small}
    \end{center}
    \vskip -0.1in
\end{table*}

\textbf{Evaluation.}
We evaluate language models across a range of widely used tasks. 
Specifically, we assess four reasoning tasks: (1) grade-school science, using \textit{ARC-easy} and \textit{ARC-challenge}~\cite{clark2018thinksolvedquestionanswering}; (2) physical commonsense, using \textit{PIQA}~\cite{piqa}; (3) social situations, using \textit{SIQA}~\cite{sap-etal-2019-social}. 
Moreover, following \cite{wang2018glue}, we evaluate language models (one-shot) on single-sentence tasks, similarity and paraphrase tasks, as well as natural language inference (NLI) tasks. 
These include: (4) \textit{SST-2}~\cite{sst2} for binary sentiment classification, (5) \textit{MRPC}~\cite{dolan-brockett-2005-automatically} for assessing semantic equivalence between sentence pairs, and (6) \textit{MNLI}~\cite{mnli} for determining entailment, contradiction, or neutrality between sentence pairs. 
The evaluation code is based on the LM Evaluation Harness~\cite{eval-harness}, and the metric used is accuracy.

\textbf{Results.}  
Table~\ref{tab:lm} presents the evaluation results, showing that \modelname\ achieves stable improvements over the vanilla Transformer in most tasks, with the highest overall average accuracy. 
While our baseline models achieve top results in several tasks, their performance is unstable, with some tasks experiencing drops in accuracy compared to the vanilla Transformer.
Meanwhile, the fluctuating impact on performance observed with Differential Attention further indicates that the additional coefficients introduced to stabilize training may not be robust enough across different training datasets or model architectures.
In contrast, \name\ does not introduce any additional modules or (un)learnable coefficients, ensuring effective training without added complexity. 

\textbf{Scalability.}
To demonstrate the scalability of \name, we train larger models with 600 million parameters on 250 billion tokens, and 1 billion parameters on 50 billion tokens.
Detailed architecture and training hyperparameters are provided in Appendix~\ref{apx:llm}. 
Due to the prohibitive cost of implementing all baselines at this scale, we only compare \modelname\ with the vanilla Transformer.
Meanwhile, training models at a larger scale or using more tokens is unfeasible for us.
As shown in Table~\ref{tab:llm}, larger \modelname s still outperforms Transformers.
Figure~\ref{fig:res} in the appendix presents performance across tasks for 600M models trained with 100, 150, and 200 billion tokens, where \modelname\ achieves better results in most comparisons.
These results highlight the potential of \name\ for pretraining advanced large language models. 

\subsection{Diffusion Models for Image Generation}

We adopt U-ViT~\cite{bao2022all} as the backbone model, which leverages the Vision Transformer~\cite{dosovitskiy2021an} architecture as a diffusion model. 
We enhance U-ViT by replacing its softmax attention (except in the first and last layers) with \name, resulting in our modified diffusion model, U-ViC. 
The remaining baselines from the previous section are not compared, as they were originally designed and evaluated only on NLP tasks.

All models have 44 million parameters.
Training is performed on the CIFAR-10 dataset~\cite{Krizhevsky2009LearningML} for unconditional image generation and the MS-COCO dataset~\cite{lin2015microsoftcococommonobjects} for text-conditioned image generation. 
For detailed information on the model architecture and training hyperparameters, please refer to Table 2 and Table 5 in~\cite{bao2022all}. 
We evaluate performance using the FID score~\cite{fid}, where lower values indicate better performance. 
All models are trained on 8$\times$A800 GPUs.
Some environment differences may lead to slightly different reproduced results.
Table~\ref{tab:fid} presents the results.
Our U-ViC model shows superior performance across both tasks.
Figure~\ref{fig:vit-sample} illustrates generated samples.

\begin{table}[t]
\vskip -0.08in
    \caption{Our U-ViC achieves lower FID in unconditional image generation (CIFAR-10) and text-to-image generation (MS-COCO).}
    \vskip 0.1in
\begin{center}
\begin{small}
    \label{tab:fid}
    \resizebox{0.75\linewidth}{!}{
    \begin{tabular}{l|c|c}
    \toprule
    Model & CIFAR-10 & MS-COCO \\\midrule
       U-ViT  & 3.39 & 5.99 \\
       U-ViC (Ours) & \textbf{3.27} & \textbf{5.85} \\\bottomrule
    \end{tabular}}
    \end{small}
    \end{center}
    \vskip -0.15in
\end{table}

\section{Discussions}
\label{sec:discuss}
We discuss some important characteristics of \name, which may help future research.

\subsection{Convergence of \modelname}
\label{sec:converge}
As briefly mentioned earlier, in a deep \modelname, we preserve the softmax attention in both the first and last layers.
This aims to maintain, or even improve, the convergence rate compared to a vanilla Transformer.

We initially attempted to apply \name\ across all layers, but this resulted in slower convergence. 
Our observations of attention dynamics revealed that, at the beginning of training, the signs of the attention weights do not accurately represent meaningful semantics due to the query-key inner product not being fully learned, which leads to diverted optimization direction. 
The model appears to learn a shortcut at first, achieving lower training loss compared to the vanilla Transformer but then becomes stuck, allowing the vanilla Transformer to surpass it as training progresses.
In contrast, softmax attention offers a near-uniform distribution that facilitates smoother initial training.
Through empirical trials, we found that preserving softmax attention in the first layer of \modelname\ largely resolves the convergence issue.
Regarding the preservation of softmax attention in the last layer, we observed that the signs of the attention weights in the last layer are highly consistent—either completely positive or completely negative. 
This suggests that the last layer may not require the expressiveness of negative weights.
The training losses for these variants are presented in Table~\ref{tab:loss} where \(\text{\modelname}_{\text{all}}\) begins to exhibit higher losses after 2,500 steps. 

\begin{figure}[!t]
\begin{center}
\centerline{\includegraphics[width=0.98\linewidth]{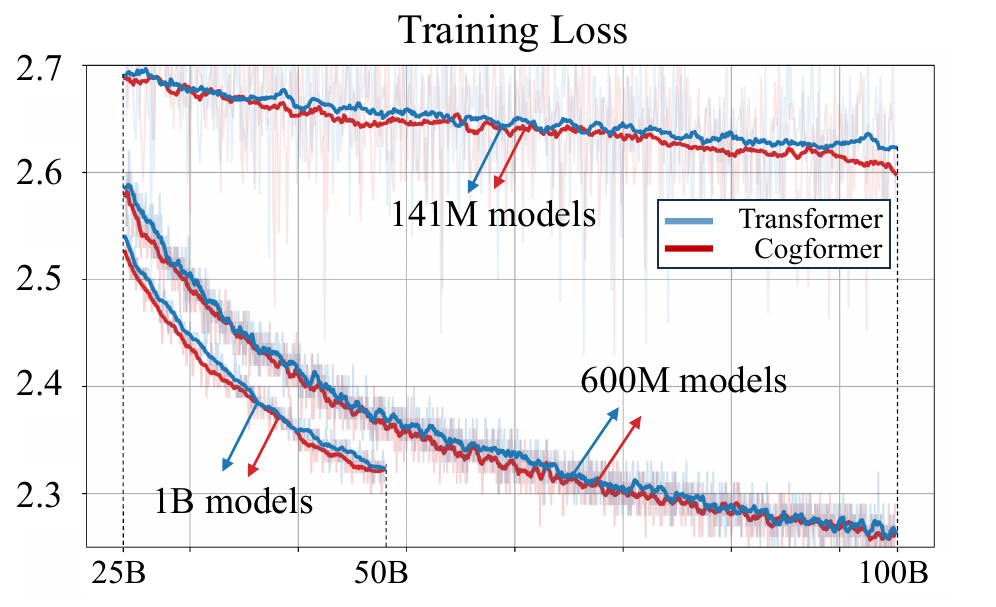}}
    \caption{Training losses for models at different scales over the first 100 billion tokens (50 billion for the 1B-parameter models).}
    \label{fig:loss}
\end{center}
\vskip -0.3in
\end{figure}

Figure~\ref{fig:loss} shows the training losses for the models in our main experiments. 
When softmax attention is applied in the first and last few layers, \modelname\ demonstrates more effective training across different scales, achieving a generally lower loss compared to the Transformer.

\begin{table}[t]
    \caption{Training loss over the first 4,500 steps for various models: $\text{\modelname}_{\text{all}}$ (using \name\ in all layers), $\text{\modelname}_{\text{0}}$ (applying softmax in layer 0), $\text{\modelname}_{\text{0,last}}$ (applying softmax in both the first and last layers), and the vanilla Transformer (141M).}
        \vskip 0.1in
\begin{center}
\begin{small}
    \label{tab:loss}
    \begin{tabular}{lccccc}
    \toprule
    \multirow{2}{*}{Model} & \multicolumn{5}{c}{Training Step} \\\cmidrule(lr){2-6}
    & 500 & 1,500 & 2,500 & 3,500 & 4,500\\\midrule
        Transformer & 6.14 & 4.74 & 3.95 & 3.53 & 3.38 \\\midrule 
       $\text{\modelname}_{\text{all}}$ & 6.07 & 4.73 & 4.03 & 3.60 & 3.45 \\
        $\text{\modelname}_{\text{0}}$ & 6.14 & 4.72 & 4.02 & 3.55 & 3.39 \\
        $\text{\modelname}_{\text{0,last}}$ & 6.15 & 4.73 & 3.99 & 3.53 & 3.38 \\\bottomrule
    \end{tabular}
    \end{small}
    \end{center}
    \vskip -0.1in
\end{table}

\subsection{\name\ Produces More Diverse Attention Patterns and Less Sink}

Our analysis reveals that \name\ generates more diverse attention patterns compared with softmax attention. 
To investigate this, we examined the attention patterns across all heads in both $\text{\modelname}_{\text{141M}}$ and $\text{Transformer}_{\text{141M}}$, using the abstract section of ``Attention Is All You Need''~\cite{NIPS2017_3f5ee243} as an input case.
A comparison of Figures~\ref{fig:attn-vanilla} and~\ref{fig:attn-my} shows that most heads in the vanilla Transformer exhibit sparse attention weights and a significant attention sink~\cite{xiao2024efficientstreaminglanguagemodels}, indicating that these heads are relatively inactive when processing the current input~\cite{off}. 
In contrast, \modelname\ displays more diverse attention patterns in its middle layers with a reduced attention sink, which suggests that more heads are engaged in processing the input. 
We hypothesize that flexible attention patterns introduced by negative weights may lead to reduced parameter redundancy, but we currently lack direct evidence to quantify this reduction. 
Additionally, the diminished attention sink could potentially enhance extrapolation capabilities~\cite{chen2024hopenovelpositionalencoding}, KV cache compression~\cite{liu2023scissorhandsexploitingpersistenceimportance}, high-context-awareness task performance~\cite{lin2024mixtureincontextexpertsenhance}, and mitigate the lost-in-the-middle issue~\cite{liu2023lostmiddlelanguagemodels}.
We will explore these questions in future research.

\section{Related Works}
Numerous efforts have modified the softmax function in the attention mechanism mainly for efficiency purposes, yet these attempts have not removed the constraints on non-negative weights. 
Some studies have proposed softmax alternatives such as ReLU~\cite{shen2023studyrelusoftmaxtransformer,wortsman2023replacingsoftmaxreluvision}, sigmoid~\cite{ramapuram2024theoryanalysisbestpractices}, cosine-based distance re-weighting~\cite{qin2022cosformerrethinkingsoftmaxattention}, and learnable activations~\cite{liu2024consmaxhardwarefriendlyalternativesoftmax}. 
The approaches discussed in the ``Baselines'' paragraph of Section~\ref{sec:exp}, which occasionally produce negative weights as a by-product of their specialized operations, showed the potential of an attention mechanism tailored to incorporate negative weights.
In this paper, we propose \name, which introduces negative weights for increased expressiveness without requiring additional parameters or meticulous hyperparameter tuning—unlike the approaches discussed in Section~\ref{sec:exp}.
A \modelname\ requires the maintenance of softmax attention in both the first and last few layers, highlighting the distinct characteristics of these layers, whose unique behavior has been discussed in several studies~\cite{gong2024mixtureofmodulesreinventingtransformersdynamic,cancedda2024spectralfiltersdarksignals}.

\section{Conclusions}
We introduce \name, a novel attention mechanism that incorporates negative weights. 
Mechanistic interpretations reveal that negative attention weights enhance the expressiveness of Transformers and increase robustness against representational collapse. 
We train \modelname, a new variant of the Transformer that integrates \name\ as its attention layer, as both language models and image generation diffusion models. 
\modelname\ outperforms Transformers on these tasks.
We also discuss several properties of \name\ that could be beneficial for various applications, which we will explore in future work.

\section*{Acknowledgements}
We appreciate the valuable discussions with Shen Nie about the U-ViT models and to Yining Qian, Jia-Nan Li, Songhao Wu, and Shuqi Li for their helpful writing suggestions.

\bibliography{main}
\bibliographystyle{icml2025}


\newpage
\appendix
\onecolumn

\section{Efficiency Analysis} 
\label{apx:time}
\begin{figure*}[h]
\begin{center}
\centerline{\includegraphics[width=0.9\linewidth]{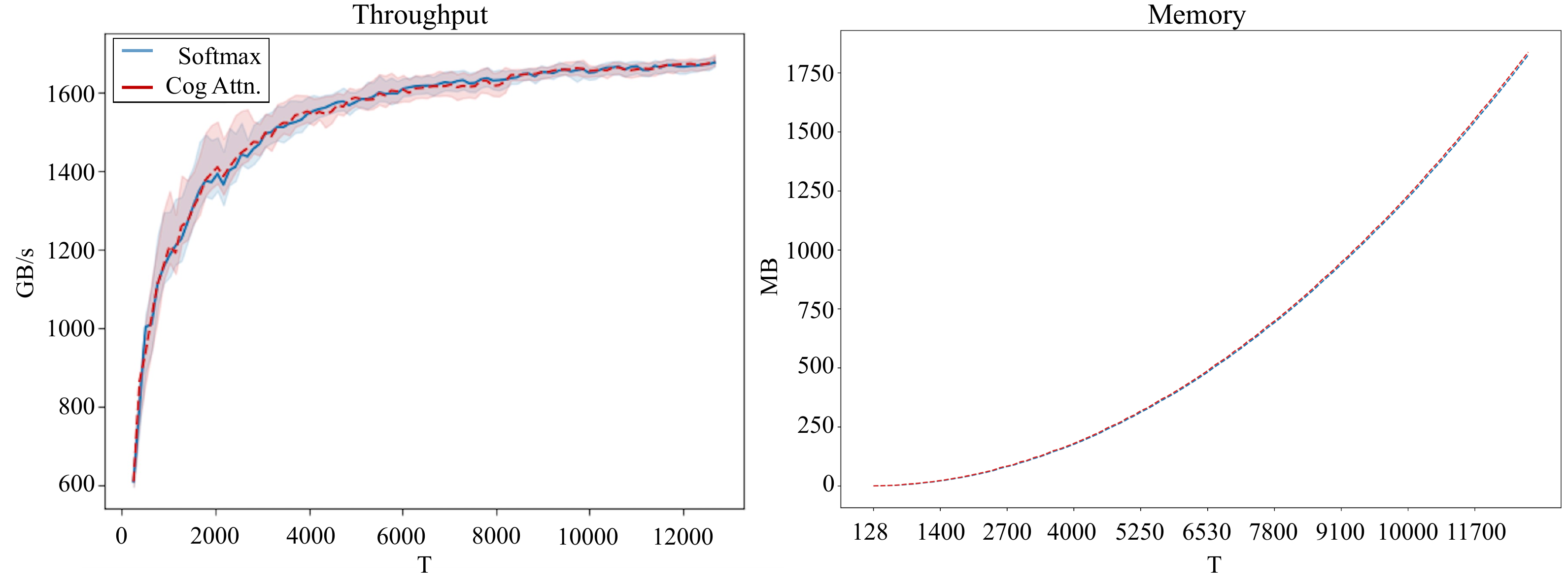}}
\caption{\name\ is as efficient as softmax attention.}
\label{fig:time}
\end{center}
\end{figure*}

We compare the efficiency of \name\ and softmax attention, both implemented using Triton fused kernels, on a single A800-80G GPU.
Figure~\ref{fig:time} illustrates the throughput and memory costs for \name\ and softmax attention, given a QK-product matrix of size \( T \times T \). 
Compared to softmax attention, \name\ introduces no additional time overhead and a  negligible increase in memory usage when $T$ is very large.

\section{How Do Negative Attention Weights Mitigate the ``Over-Squashing'' Issue}
\label{apx:proof}

\begin{table}[h]
    \caption{An overview of symbols used in the theorem for convenience.}
        \vskip 0.1in
\begin{center}
\begin{small}
    \begin{tabular}{c|c}
    \toprule
    Symbol & Explanation\\\midrule
        $L$ & Maximum number of layers in the model. \\
        $l$ & Index of a layer, ranging from $0$ to $L$; layer $0$ represents the embedding layer. \\
       $\textbf{v}^{l}_{i}$  &  The $i$-th input at the $l$-th layer. \\
        $\textbf{y}_{i}$ & The $i$-th output of the entire model. \\
        $\alpha^{l}_{j,i}$ & Attention weight from position $j$ to position $i$ at layer $l$. \\
        $\sigma$ & The maximum Lipschitz constant of any MLP layer. \\
        $\sigma^{l}$ & Lipschitz constant of the MLP at layer $l$. \\
        $\beta^{l}_{i}$ & The effect of $i$-th normalization module at layer $l$, where $i\in \{1,2\}$ \\
        $\beta_{3}$ & The effect of normalization module applied after all layers.\\
        $k_l$ & Enumeration index along the sequence length dimension within layer $l$.\\\bottomrule
    \end{tabular}
    \label{tab:my_label}
    \end{small}
    \end{center}
\end{table}

We explain \modelname's enhanced robustness against representational collapse using the theorem B.5 from \cite{barbero2024transformers}. 
The theorem states that for an input sequence \(\mathbf{v}^{0}_{1}, \dots, \mathbf{v}^{0}_{n}\), the following inequality holds:
\[
\lVert \frac{\partial \mathbf{y}_{n}}{\partial \mathbf{v}^{0}_{i}} \rVert \leq C \sum_{k_{1} \geq i} \dots \sum_{k_{L} \geq k_{L-1}} \bar{\alpha}^{L-1}_{n,k_L} \prod_{l=2}^{L-1} \bar{\alpha}^{l-1}_{k_l,k_{l-1}} \bar{\alpha}^{0}_{k_1,i},
\]
where \(\bar{\alpha}^{l}_{j,i} = \lvert \frac{\alpha^{l}_{j,i}}{\beta^{l}_{1}} + \mathbbm{1}(i=j) \rvert \) and \(C = \frac{1}{\beta_{3}} \prod_{l=1}^{L} \left( \frac{\sigma}{\beta^{l}_{2}} + 1 \right)\).

We can see that allowing attention weights to be negative reduces the upper bound of \(\left\| \frac{\partial \mathbf{y}_{n}}{\partial \mathbf{v}^{0}_{i}} \right\|\). 
Specifically, for any layer \(l\), when \(k_{l} = k_{l-1}\), the expectation of \(\bar{\alpha}^{l}_{k_{l},k_{l-1}}\) decreases, thereby lowering the right side of the inequality. 
An intuitive explanation is that negative weights can adaptively diminish the information flow from position \(k_{l-1}\) in layer \(l-1\) to the same position \(k_{l-1}\) in layer \(l\). 
Consequently, for any \(k_{l+1} \geq k_{l-1}\), \(\mathbf{v}^{l}_{k_{l+1}}\) is less likely to be over-squashed by \(\mathbf{v}^{l}_{k_{l-1}}\).
By ``adaptively diminish,'' we mean that when \(\mathbf{v}^{0}_{i}\) contains irrelevant information (e.g., token \(i\) is ``1'' while the query is ``To find a zero''), \modelname's attention to token $i$ tends to be negative, mitigating the over-squashing of irrelevant information. 
This enables the model to focus on useful tokens.

It is important to note that the information in \(\mathbf{v}^{0}_{i}\) is not only semantic but also positional. For tasks where tokens are semantically similar (e.g., task 2 in Section~\ref{sec:rep-exp}), the position information from earlier tokens is less over-squashed into the outputs, allowing more information pathways for diverse positions.
For reference, several works have studied the critical role of positional information in Transformers' counting ability~\cite{barbero2024transformers, yehudai2024transformerscountn, golkar2024contextualcountingmechanisticstudy}.

\section{Implementation Specifics and Hyper-parameters for Larger Models}  
\label{apx:llm}  

The architectures of our middle and large models share the same position embedding, FFN structure, and vocabulary as the small language models (141M). 
The training context length is set to 4,096 tokens.  

The training hyper-parameters follow those outlined in~\cite{StableLM-3B-4E1T}: a batch size of 4 million tokens, an initial learning rate of \(3.2 \times 10^{-4}\), and a linear warmup over the first 4,800 steps, followed by a cosine decay schedule that reduces the learning rate to 4\% of its peak value. 
The AdamW optimizer~\cite{adamw} is used with \((\beta_1, \beta_2) = (0.9, 0.95)\), a gradient norm clipping threshold of 1, and a weight decay of 0.1.

A 600M-parameter model consists of 16 layers, with 24 attention heads per layer.
The hidden state dimension is 1,536, and the intermediate dimension of the MLP layers is 4,608. 
The first and last two layers use softmax attention.
Training a model (with 250B tokens) at this scale requires approximately 3,600 GPU hours (A800-80G).

A 1B-parameter model consists of 22 layers, with 28 attention heads per layer.
The hidden state dimension is 1,792, and the intermediate dimension of the MLP layers is 5,376. 
The first and last three layers use softmax attention.
Training a model (with 50B tokens) at this scale requires approximately 1,600 GPU hours (A800-80G).

\newpage
\begin{figure}[t]
\begin{center}
\centerline{\includegraphics[width=\linewidth]{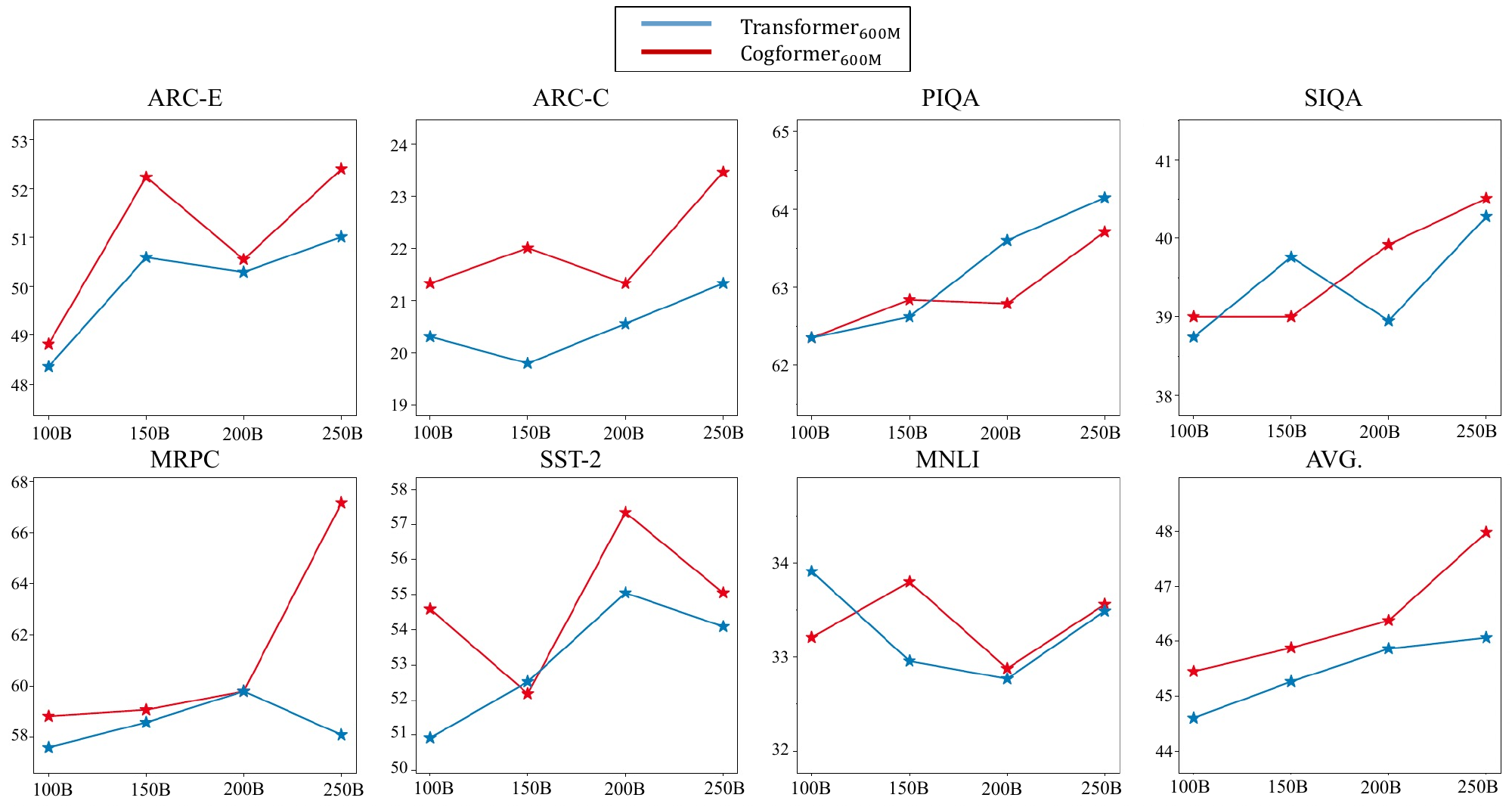}}
    \caption{Performance across tasks for models trained with 100, 150, and 200 billion tokens.}
    \label{fig:res}
\end{center}
\end{figure}

\begin{figure*}[t]
\begin{center}
\centerline{\includegraphics[width=0.65\linewidth]{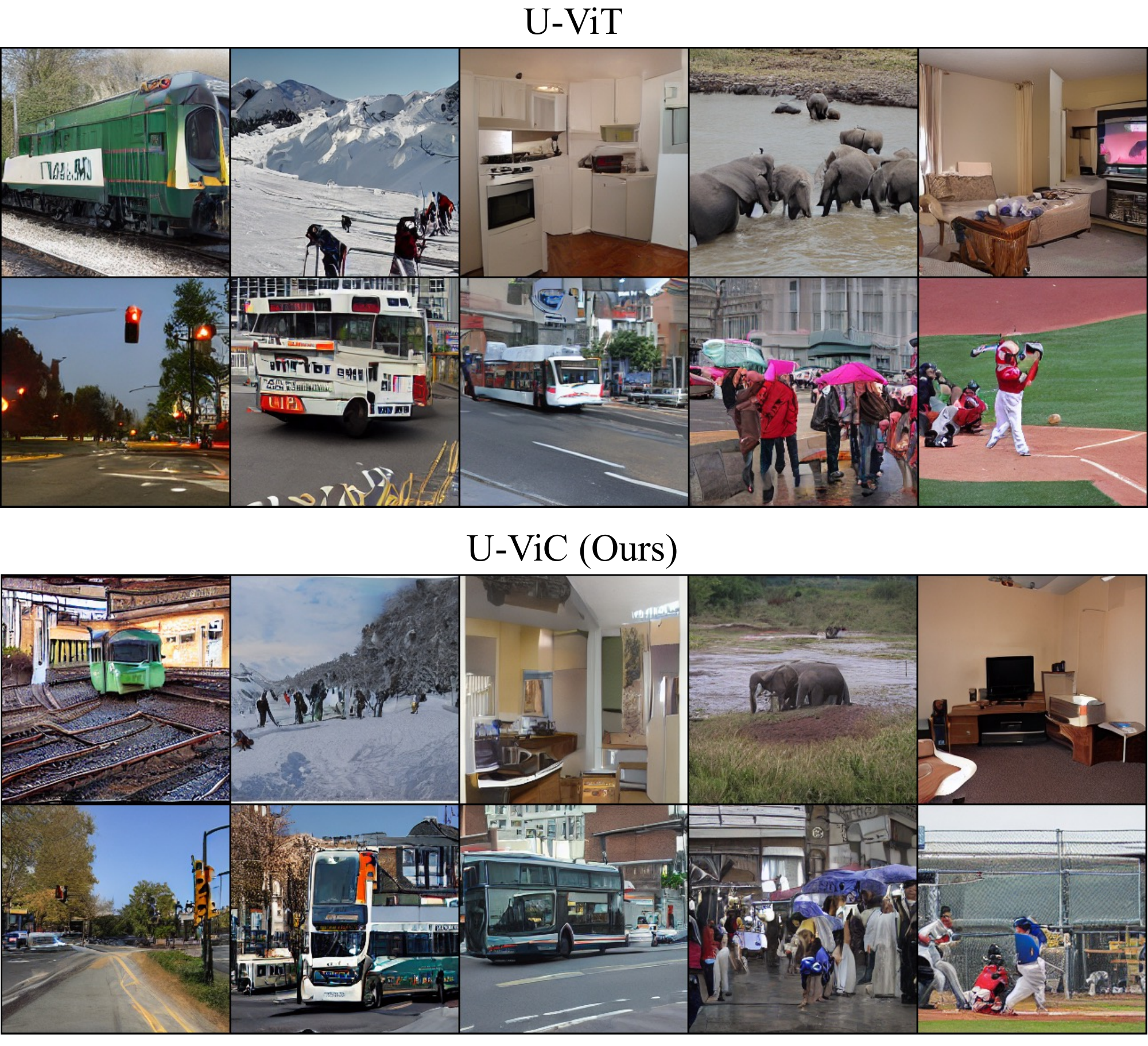}}
    \caption{Text-to-image generation samples on MS-COCO. 
    The figures are generated by the best checkpoint with lowest validation FID, and not cherry-picked.}
    \label{fig:vit-sample}
\end{center}
\end{figure*}

\begin{figure*}
    \centering
    \includegraphics[width=\linewidth]{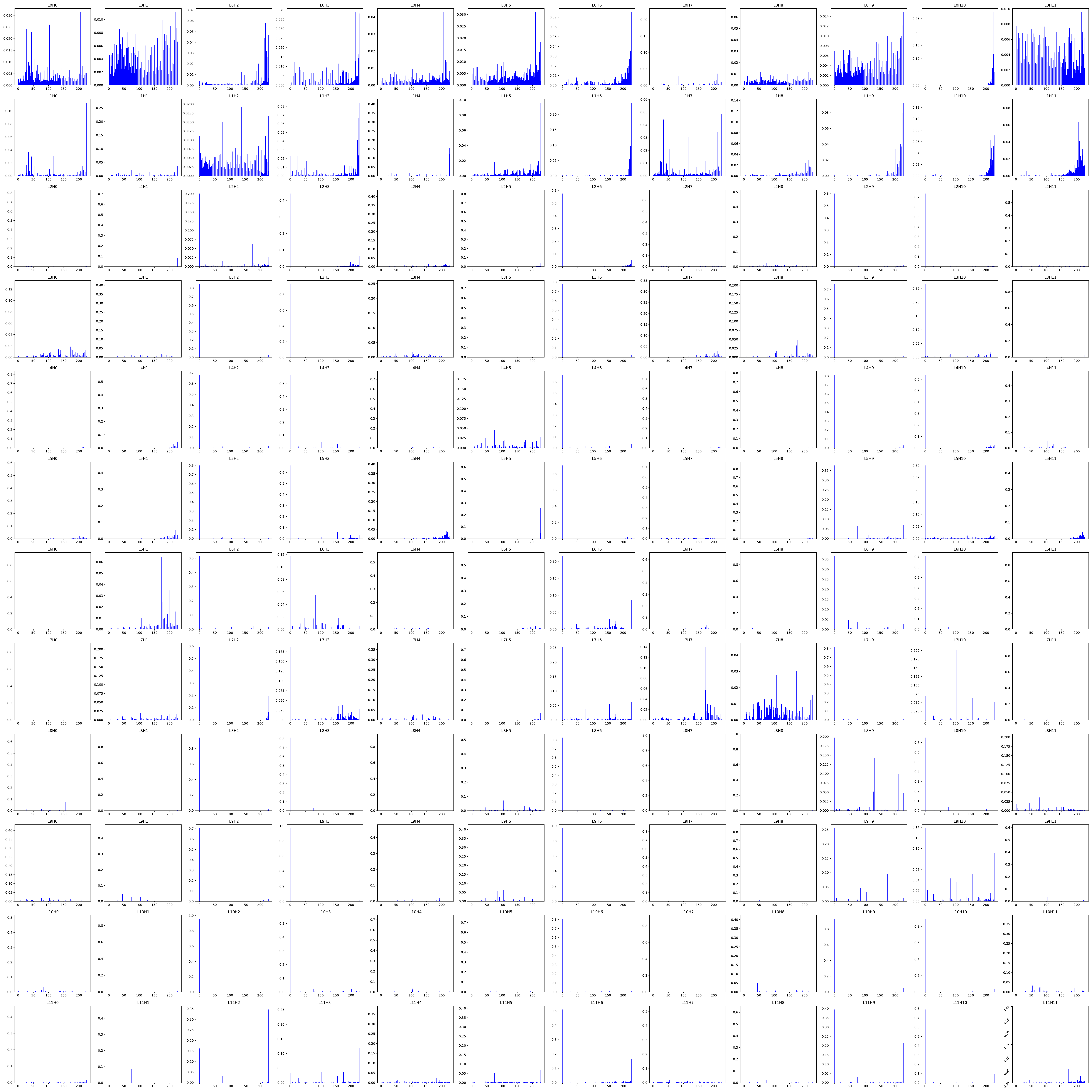}
    \caption{Attention patterns obtained from Transformer.}
    \label{fig:attn-vanilla}
\end{figure*}

\begin{figure*}
    \centering
    \includegraphics[width=\linewidth]{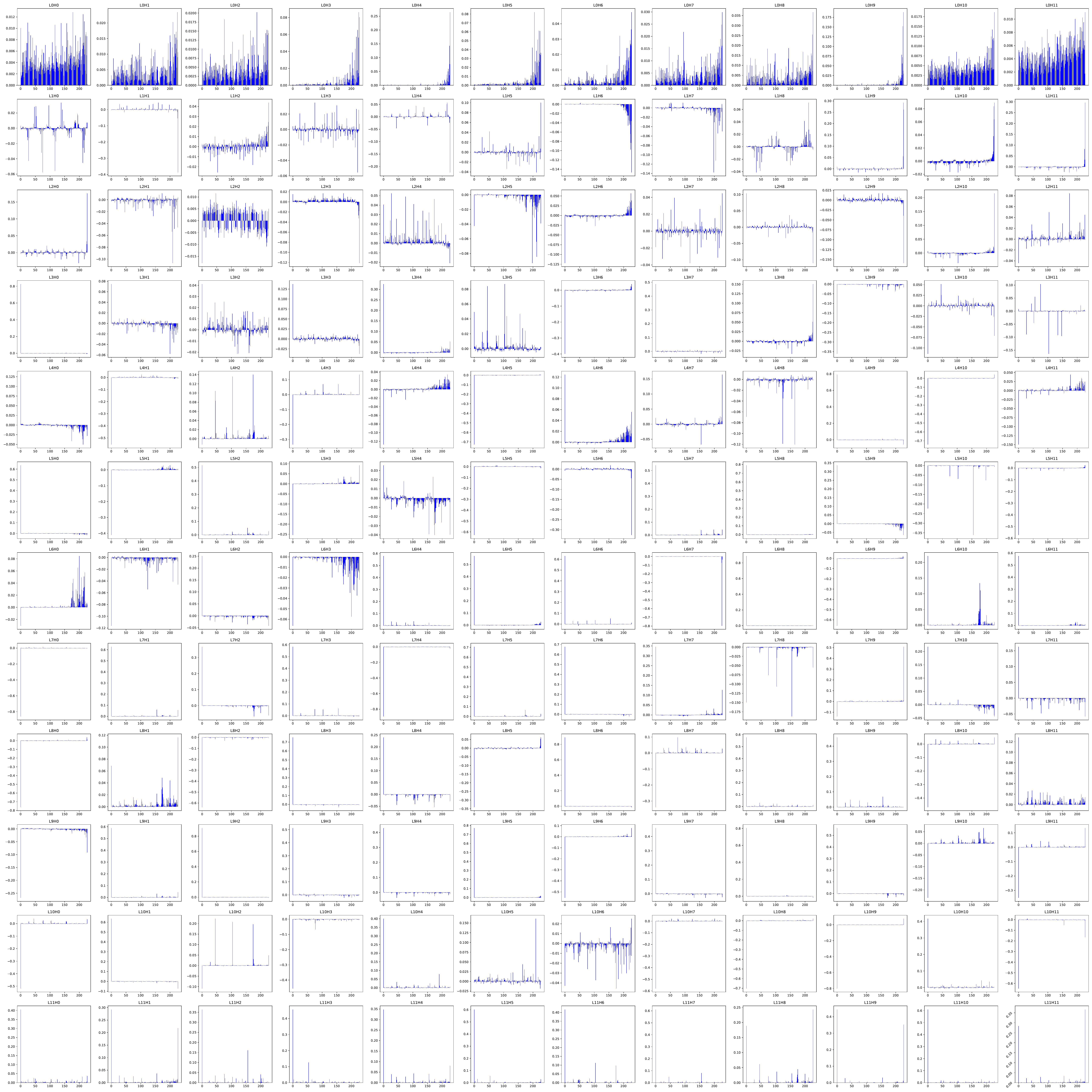}
    \caption{Attention patterns obtained from \modelname. }
    \label{fig:attn-my}
\end{figure*}

\end{document}